\documentclass{article}
\usepackage{spconf,amsmath,graphicx}

\def\BigRoman{\uppercase\expandafter{\romannumeral\number\count 255 }}
\def\Romannumeral{\afterassignment\BigRoman\count255=}
\usepackage{multirow}
\usepackage{color, colortbl}
\usepackage[justification=centerlast]{caption}
\definecolor{LightGray}{gray}{0.9}
\definecolor{DarkGray}{gray}{0.75}
\usepackage{amsmath}
\usepackage{amssymb}
\usepackage{enumerate}
\usepackage[ ]{algorithm2e}
\SetKwInput{KwInput}{Input}
\SetKwInput{KwOutput}{Output}
\makeatletter
\newcommand{\nosemic}{\renewcommand{\@endalgocfline}{\relax}}
\usepackage{booktabs}
\usepackage{hyperref}
\newcommand{\head}[1]{\textnormal{\textbf{#1}}}
\newcommand\norm[1]{\left\lVert#1\right\rVert}

\title{Recognizing Dynamic Scenes \\
with Deep Dual Descriptor \\
based on Key Frames and Key Segments}
\name{Sungeun Hong, Jongbin Ryu, Woobin Im, Hyun S. Yang }
\address{School of Computing, Korea Advanced Institute of Science and Technology, Republic of Korea}

\begin{document}
%
\maketitle
\begin{abstract}
Recognizing dynamic scenes is one of the fundamental problems in scene understanding, which categorizes moving scenes such as a forest fire, landslide, or avalanche.
While existing methods focus on reliable capturing of static and dynamic information, few works have explored frame selection from a dynamic scene sequence. 
In this paper, we propose  dynamic scene recognition using a deep dual descriptor based on `key frames' and `key segments.' Key frames that reflect the feature distribution of the sequence with a small number are used for capturing salient static appearances. Key segments, which are captured from the area around each key frame, provide an additional discriminative power by dynamic patterns within short time intervals. To this end, two types of transferred convolutional neural network features are used in our approach. A fully connected layer is used to select the key frames and key segments, while the convolutional layer is used to describe them. We conducted experiments using public datasets as well as a new dataset comprised of 23 dynamic scene classes with 10 videos per class. The evaluation results demonstrated the state-of-the-art performance of the proposed method.

\end{abstract}
\begin{keywords}
 dynamic scene classification, key frame, key segment,  convolutional neural networks, transfer learning, deep dual descriptor (D3)
\end{keywords}

\section{INTRODUCTION}
Dynamic scene recognition is an increasingly important problem in identifying a moving scene, such as a forest fire, waterfall, or avalanche \cite{du2015dynamic}. A major challenge in dynamic scene recognition is the reliable capturing of spatial and temporal information from dynamic scenes. Over the past decade, considerable efforts \cite{chan2007classifying, ghanem2010maximum,  xia2012compact, shroff2010moving, derpanis2012dynamic} have been devoted to these issues. The various approaches can be divided into two categories: simultaneous modeling and separate modeling. In simultaneous modeling, numerous spatiotemporal descriptors are constructed. Each one describes a small 3D (x-y-t) cuboid. In separate modeling, the spatial and temporal features are separately modeled then combined, i.e., late fusion.

Early investigations of simultaneous modeling have involved linear dynamical systems (LDS) \cite{doretto2003dynamic}, which analyze video sequences in the Stiefel manifold. This approach consists of a hidden-state process that encodes motions from the video as well as the appearance of each frame, which is conditioned on the current hidden state. Several subsequent approaches \cite{chan2007classifying, ghanem2010maximum,  xia2012compact} considered the LDS model to represent dynamic texture that closely relates to a dynamic scene. However, these approaches have not shown promising results in dynamic scene recognition because the first-order Markov property or linearity assumption of LDS is restrictive in dynamic scenes \cite{shroff2010moving}. 

To better represent dynamic scenes, Derpanis et al. \cite{derpanis2012dynamic} introduced multi-scale orientation features using 3D Gaussian third-derivative filters. The bag of features (BoF) scheme \cite{csurka2004visual} was additionally applied to represent several spatiotemporal patches in dynamic scenes \cite{feichtenhofer2014bags, feichtenhofer2016dynamic}. Encouraged by the promising results of convolutional neural networks (CNNs) \cite{krizhevsky2012imagenet, girshick2014rich, ji20133d}, Tran et al. \cite{tran2015learning} recently proposed a convolutional three-dimensional (C3D) architecture that is a spatiotemporal version of CNN. The performances of CNN-based features, i.e., C3D, were shown to be significantly better than those of hand-crafted features in public dynamic scene datasets.


On the other hand, separate modeling approaches were advanced with the introduction of the methods in \cite{shroff2010moving}. To capture spatial information, the authors first extracted GIST features \cite{oliva2001modeling} from whole frames in the sequence. For temporal information, chaotic features \cite{shroff2010moving} are calculated and then combined with spatial GIST features for video-level representation. 
Meanwhile, Yang et al. \cite{yang2016dynamic} suggested an ensemble scheme for aggregating spatial and temporal features from dynamic scenes. Tremendous spatial features, such as local binary pattern (LBP) \cite{ojala2002multiresolution}, Gabor \cite{grigorescu2002comparison}, and GIST \cite{oliva2001modeling}, are extracted from each frame in the sequence, and naive dynamic features \cite{doretto2003dynamic} are used for temporal information. 
Furthermore, Qi et al. \cite{qi2016dynamic} suggested a transferred CNN feature (TCoF)-based approach. A spatial feature is extracted for a chosen number of frames by a pre-trained CNN. To capture spatial information, each frame is used as the input of the pre-trained CNN, while the differences between two adjacent frames are used as the inputs to capture temporal information.

Overall, recent experiments have shown that separate modeling, including tremendous hand-crafted spatial features \cite{yang2016dynamic} or transferred CNN features \cite{qi2016dynamic}, outperform methods involving simultaneous modeling \cite{derpanis2012dynamic,  feichtenhofer2014bags, feichtenhofer2016dynamic,  theriault2013dynamic, feichtenhofer2013spacetime}. Nevertheless, we believe that performance can still be improved of separate modeling of dynamic scenes in two aspects as outlined below.

\begin{enumerate}[$\bullet$] \item
\textbf{Frame selection}: There are several frame selection strategies for capturing the static appearances of dynamic scenes. Shroff et al. \cite{shroff2010moving} used whole frames in the sequence, which is a time-consuming approach that produces redundant information. Recently, randomly selected frames \cite{yang2016dynamic} or partially consecutive frames \cite{qi2016dynamic}, e.g., the first $\frac{N}{8}$ frames, were used to represent dynamic scenes. Although these approaches address computational time and redundancy issues, they do not well reflect the characteristics of a given sequence. 
\end{enumerate}

\begin{enumerate}[$\bullet$] \item	
\textbf{Temporal modeling}: In addition to the static appearance, temporal information may provide additional discriminant features. In the TCoF-based method \cite{yang2016dynamic}, the difference of two adjacent frames is used as input for the pre-trained CNN model for temporal modeling. However, the information between only two adjacent images is considered insufficient for capturing irregular motions of dynamic scenes. 
\end{enumerate}

\begin{figure}[]
	\centering
	\includegraphics[width=1.01\linewidth]{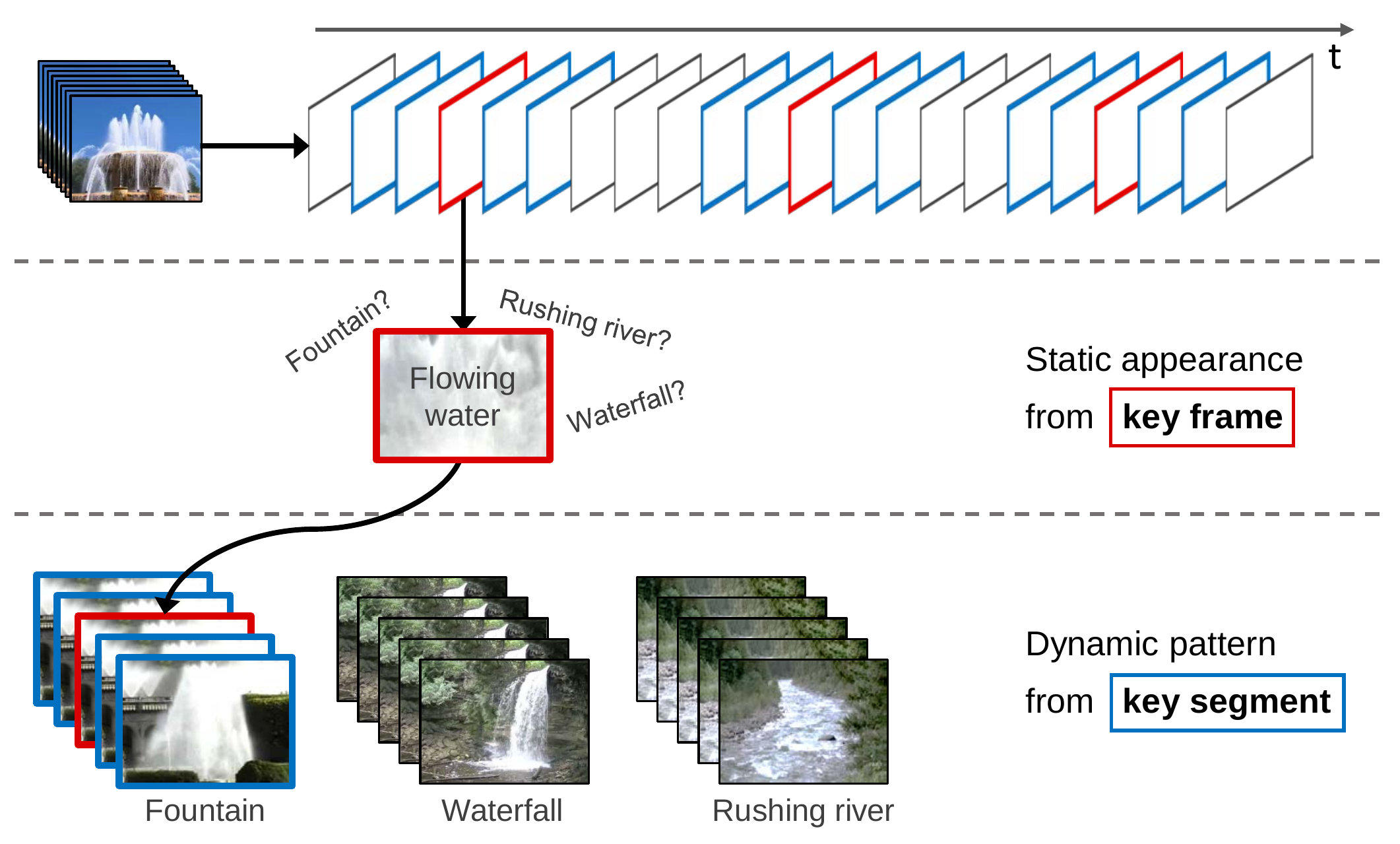}
	\caption{
	Dynamic scene representation using key frames and key segments. A key frame is used for capturing the salient static appearance of a dynamic scene. A key segment, which is captured from the area around each key frame, provides crucial discriminative power with dynamic patterns.}
	\label{fig:concept}
\end{figure}

To address the above issues, we propose a dynamic scene recognition method based on the `key frame' and `key segment.' The main concept of the proposed method is illustrated in Fig.~\ref{fig:concept}. A key frame solves the limitation of conventional frame selection, while a key segment, which is captured from the area around each key frame, efficiently provides temporal information. To this end, two types of transferred CNN features are used in our approach. A fully connected (FC) layer, which reflects the overall structure of each scene, is used to select the key frame and key segment. On the other hand, a convolutional (Conv) layer, which considers the local texture of each scene, is used to describe the key frame and key segment.

To validate the effectiveness of the proposed method, we conducted extensive experiments on both dynamic scene datasets---Maryland \cite{shroff2010moving} and Yupenn \cite{derpanis2012dynamic}---and dynamic texture datasets---Dyntex-Alpha, Dyntex-Beta, and Dyntex-Gamma \cite{peteri2010dyntex}. Currently, there exist only two benchmark datasets for dynamic scenes; thus, we constructed a dataset comprised of 23 dynamic scene classes with ten videos per class by combining the Maryland and Yupenn datasets.

The main contributions of this study are summarized as follows:
({\romannumeral 1}) We propose a CNN-based dynamic scene representation method based on key frames and key segments, which capture characteristics of a dynamic scene sequence using only a small number of frames.
({\romannumeral 2}) We investigated various factors that influence the performance of our approach, such as the frame selection strategy, number of frames, and layers for feature description.
({\romannumeral 3}) Owing to the lack of public dynamic scene datasets, we constructed a rich dynamic scene dataset of 23 classes with 10 videos per class.

\begin{figure}[b]
	\centering
	\includegraphics[width=1.01\linewidth]{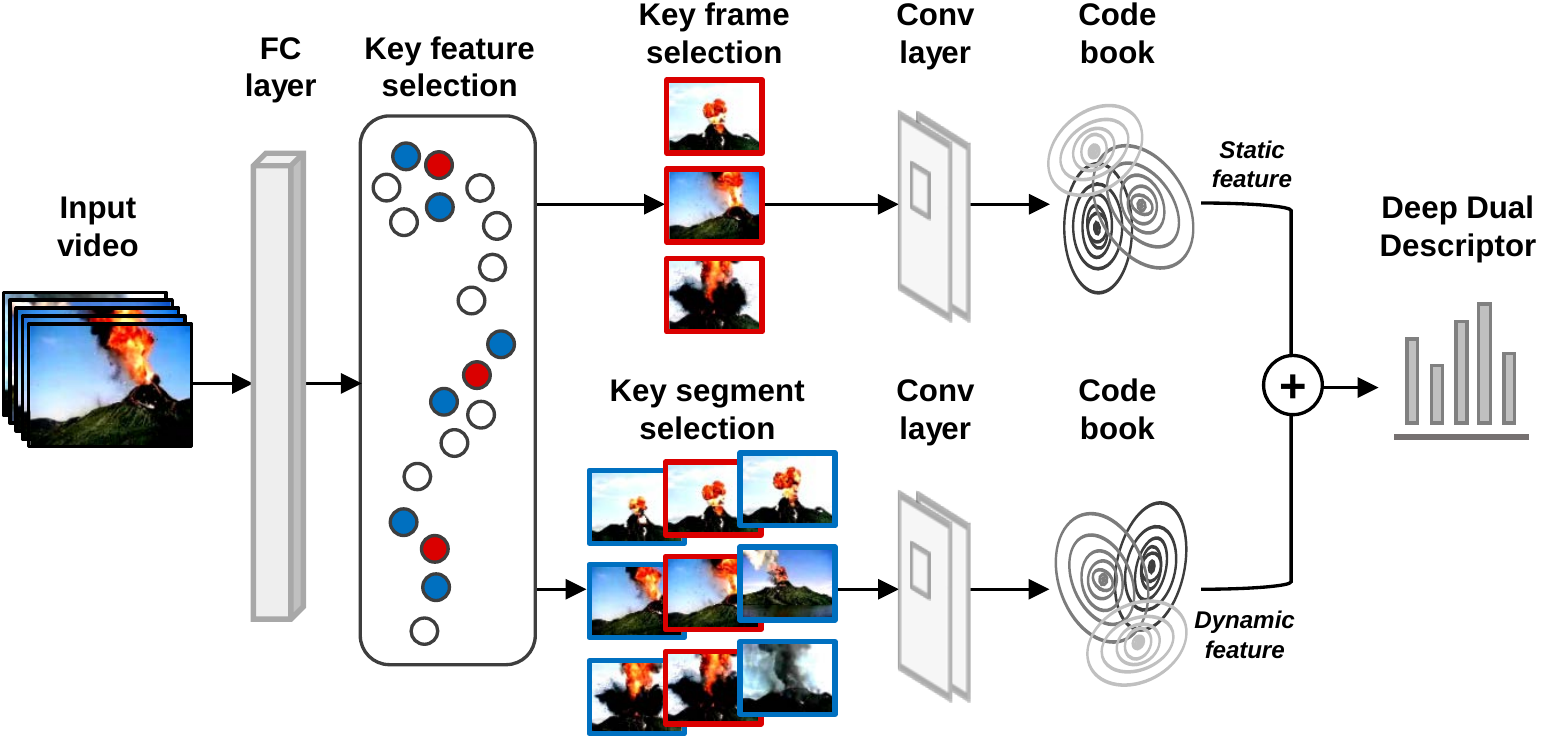}
	\caption{
	Outline of the proposed method. Two main streams exist using the Conv layer after key feature selection via the FC layer: a key frame description for the static appearance, and a key segment description for the dynamic pattern.	
		}
	\label{fig:diagram}
\end{figure}

\section{Deep Dual Descriptor}   
\label{proposed_method}

Fig.~\ref{fig:diagram} depicts a flow diagram of the proposed method. Given a dynamic scene video, we first extract features from each frame via the FC layer of the pre-trained CNN. Once all FC-features are obtained, we find key features that reflect the feature distribution of the sequence using a small number. We then select the key frames and key segments according to the key features. To describe the key frame and key segment, we use the Conv layer as a feature extractor. Static features of key frames and dynamic features of key segments are then generated by codebook learning. Finally, we combine the static and dynamic features to construct our deep dual descriptor (D3).

\subsection{Key Frame and Key Segment Selection} 
\label{Key_Segment_Selection}

Because each key segment is selected based on the temporal location of the key frame, we first must select key frames. Unlike traditional frame selection approaches, the proposed frame selection process reflects the feature distribution of a dynamic scene sequence. Inspired by partitioning around medoids (PAM) \cite{kaufman2009finding}, we first find key features that well reflect the feature distribution of whole frames. Then, each frame corresponding to each key feature is determined as a key frame. Our approach minimizes the sum of pairwise dissimilarities between features and represents the cluster center by existing features. It is, therefore, robust to noisy or outlier frames.

\begin{figure}
	\centering
	\includegraphics[width=1.015\linewidth]{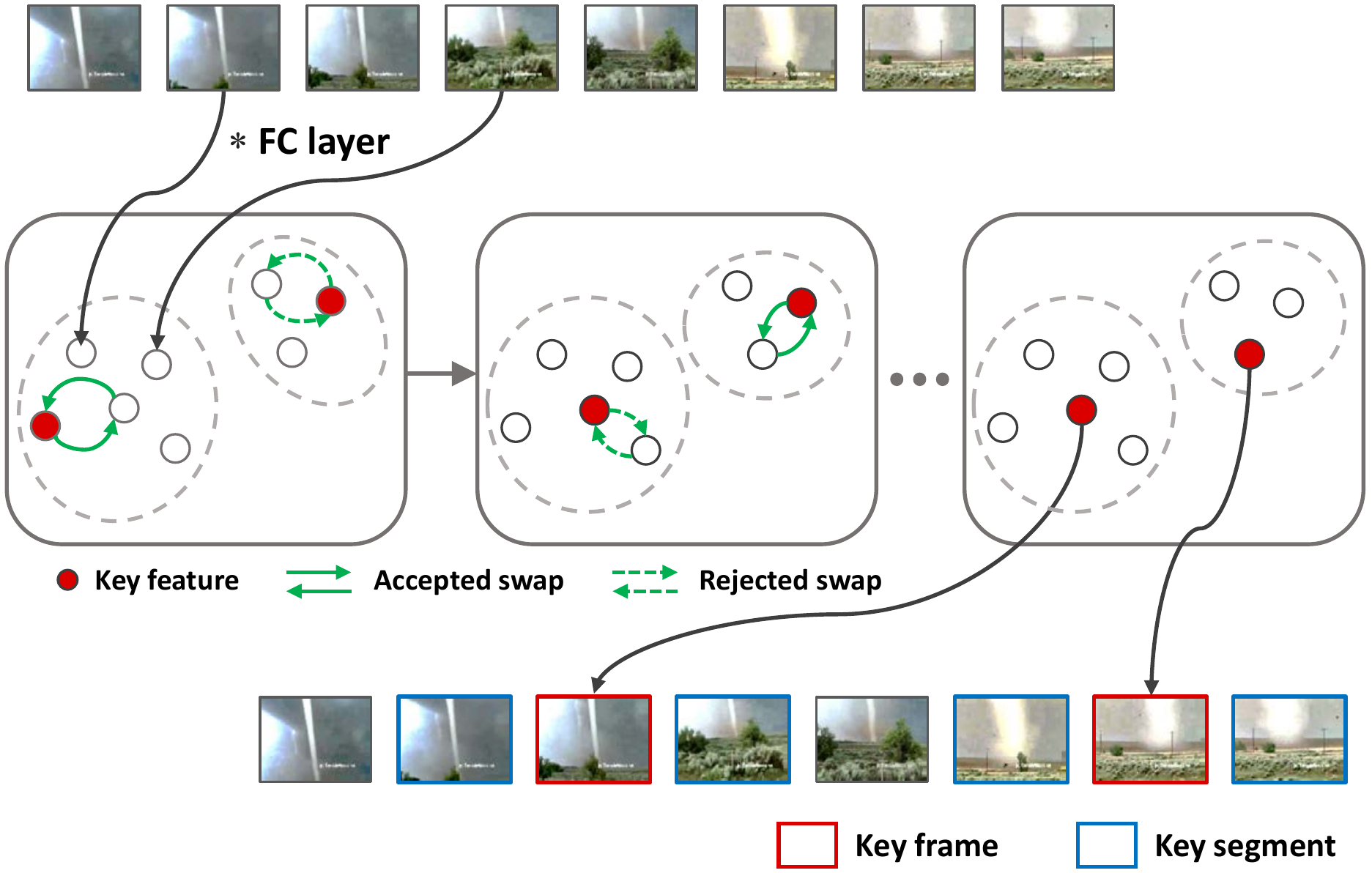}
	\caption{
	The main concept of key frame and key segment selection. Given features from whole frames, the initial key features are randomly determined. A swap process is then performed iteratively to find optimal key features. Finally, the key frame and key segment are selected according to the key feature. 	
	}
	\vspace*{-0.17in}
	\label{fig:keyfeature}
\end{figure}

The main concept of the proposed method is illustrated in Fig.~\ref{fig:keyfeature}. Given a dynamic scene sequence, we first extract spatial features from each frame via the FC layer. The circles in the figure depict corresponding FC features. Once all features are obtained, we select the initial random features, as shown in the left side of the middle row of Fig.~\ref{fig:keyfeature}. The non-key features are then clustered based on the closest key features. Once the initial clusters are generated based on the key features, each key feature is swapped with a non-key feature that belongs to its cluster. During the swap process, we verify the change of our objective function  (\ref{eq:repr}), where $K$ denotes the key feature set and $C_i$   indicates the non-key feature set belonging to the  $i$-th cluster.

\vspace*{-0.1in}
\begin{equation}
	\begin{split}
		 J = \sum_{i}\sum_{j} \norm{f_i - f_j}_2^2	\:\:\: 
		 \\\forall i: f_i \in K \quad \forall j: f_j \in C_i
		\label{eq:repr}
	\end{split}
\end{equation}

In the case of a swap that makes the objective function (i.e., the sum of the dissimilarities between the key features and non-key features) smaller than before swapping, the swap is accepted. Conversely, the swap that makes an objective function larger is rejected. If the swap is accepted, non-key features are re-clustered. This swapping process is repeated multiple times until objective function (\ref{eq:repr}) does not change in a new iteration. Overall, this approach obtains the optimized key feature set that reflects the feature distribution of all frames.

In key feature selection, calculating the pairwise dissimilarity over all features requires a considerable amount of computation because the number of FC features in a sequence is usually more than 700, and the dimension of each feature is 4,096. Therefore, to find key features, we only perform the swap process in a random subset of FC features. This procedure is repeated multiple times and clustering with the lowest objective function is retained. Finally, the frames corresponding to the key features are determined as key frames. In addition, the frames around each key frame within a short time interval are chosen as key segments.

\subsection{Feature Description} 
\label{Key_Segment_Description}
Feature description for dynamic scenes consists of three main steps: low-level feature extraction, feature aggregation for static and dynamic features, and late fusion.
\subsubsection{Low-level Feature Description} 
\label{Low-level_Feature}
In feature description, there are two separate streams, each with a different purpose: a static appearance from a key frame, and a dynamic pattern from a key segment. The first step for describing a static appearance is feature description for each key frame using the Conv layer. An output of the Conv layer is a  7$\times$7$\times$512-dimensional vector. We handle this Conv-feature as densely extracted   49 (7$\times$7) local features of  512 dimensions. It is assumed that the number of key frames is  $R$. From  $R$ key frames, we consequently obtain 49 $\times R$  static features termed as $S_{Conv} = \{c_1^1, c_1^2 \cdots c_1^{49} \cdots c_R^1, c_R^2 \cdots  c_R^{49}\}$.

For temporal modeling, rather than using temporal information of all frames, as in \cite{shroff2010moving, yang2016dynamic}, only a dynamic pattern within a short time interval is considered in our approach. Specifically, we use central moments of Conv features extracted from each frame in a key segment as dynamic features:

\vspace*{-0.1in}
\begin{equation}
	\begin{split}
	  t_i^j = \frac{1}{\tau+1} \sum_w {\left(_wc_i^j - \mu_i^j\right)}^2 
	   \quad where \:\:\mu_i^j = \frac{1}{\tau+1} \sum_w {_wc_i^j} \quad
	   \\ i = 1,2 \cdots  R\quad  j = 1,2 \cdots 49  \quad w = -\frac{\tau}{2} \cdots  {-1},\:0,\:1 \cdots  \frac{\tau}{2} \quad
	    \label{eq:var}
	\end{split}
\end{equation}
where  $i$  refers to the frame index,   $j$  indicates the index of the local region, and $w$  denotes the relative position from the middle frame of the key segment, i.e., the key frame. For instance,  $_{1}c_6^2$  denotes the second Conv feature extracted from the seventh frame. The temporal size of the key segment, which is usually below eight in our approach, is denoted by  $\tau$. As a result, we obtain $49 \times R$  dynamic features, $T_{Conv} = \{t_1^1, t_1^2 \cdots t_1^{49} \cdots t_R^1, t_R^2 \cdots  t_R^{49}\}$.

\subsubsection{Feature Aggregation and Late Fusion} 
Once low-level static feature $S_{Conv}$  and dynamic feature  $T_{Conv}$ are calculated from a video sequence, we respectively aggregate these features into a single feature vector. Three well-known feature aggregation techniques based on codebook learning are applied in our approach: bag of features (BoF) \cite{csurka2004visual}, vector of locally aggregated descriptors (VLAD) \cite{jegou2010aggregating}, and Fisher vectors (FV) \cite{perronnin2010improving}.

Here, we describe the details of feature aggregation in the case of  $S_{Conv}$. Nonetheless, all procedures are likewise applied to   $T_{Conv}$. Assume that a codebook with  $k$   visual words, $C = [ v_1 , v_2\;\; \ldots \;\; v_k]$, has been learned from the training set, and $S_{Conv}$  is given from a query video. For convenience, because the frame index within a sequence is meaningless in our feature aggregation, we change the notation of static features from $S_{Conv} = \{c_1^1, c_1^2 \cdots c_1^{49} \cdots c_R^1, c_R^2 \cdots  c_R^{49}\}$ to   $S_{Conv} = \{c_1, c_2 \cdots c_{49 \times R}\}$.

According to the method of assigning each visual word to the given local features, feature aggregation approaches are divided into two types: hard assignment and soft assignment. BoF and VLAD representations are based on the hard assignment. Once codebook  $C$ and feature set  $S_{Conv}$ are given, each feature in a BoF representation is first assigned to the closest visual word. The histogram of each visual word is then used as a feature vector. VLAD is similar to BoF; however, instead of using a histogram, the differences between each vector and the nearest visual word are accumulated as follows: 

\vspace*{-0.1in}
\begin{equation}
	\begin{split}
		 \Psi_j = \sum_{\forall i: NN(c_i)=j} c_i - v_j \qquad\qquad\quad \\ where\:NN(c_i) = \underset{j}  {\operatorname{argmin}} \norm{c_i - v_j}_2 \quad\quad\:\:
		 \\i=1, 2\:\ldots \:49 \times R \qquad j=1, 2\: \ldots \:k \quad\quad
		 \label{eq:vlad}
	\end{split}
\end{equation}

Only a single visual word is assigned to local features; therefore, the relevance of the local features to other visual words can be lost in BoF and VLAD. Following the work in \cite{cimpoi2015deep}, we use the soft-assignment FV, which encodes both first- and second-order statistics between each local feature and a Gaussian mixture model (GMM):

\vspace*{-0.1in}
\begin{equation}
	\begin{split}
	 \qquad \Phi_j^{(1)} = \frac{1}{49R\sqrt{w_k}} \sum_{i=1}^{49R}  \alpha_i(j)\left({\frac{c_i-\mu_j}{\sigma_j}}\right) \qquad\:\:
	 \\\Phi_j^{(2)} = \frac{1}{49R\sqrt{2w_k}} \sum_{i=1}^{49R} \alpha_i(j)\left({\frac{\left(c_i-\mu_j\right)}{\sigma_j^2}^2-1}\right) \quad
	 \\  i=1, 2\:\ldots \:49 \times R \qquad j=1, 2\: \ldots \:k \qquad\quad\;
		 \label{eq:fv}
	\end{split}
\end{equation}
where $w_j$, $\mu_j$, and $\sigma_j$  denote the mixture weights, means, and diagonal covariances of the $j$-th Gaussian, respectively. Moreover, $\alpha_i(j)$   is the soft assignment weight of the  $i$-th feature  $c_i$ to the $j$-th Gaussian. From results of comparative experiments, FV showed the best performance among feature aggregation approaches. Thus, we used FV representation in the remainder of our experiments.

We furthermore applied feature aggregation to dynamic features with the same approach that was used for the static features. Finally, we concatenate the FV from static features and FV from dynamic features and denote the result as the deep dual descriptor. In the remainder of this paper, we denote FV from static features as  D3$_s$, FV from dynamic features as  D3$_d$, and combined version (static \texttt{+} dynamic)  as  D3.

\section{Experimental Results}
\label{experiment}
In Section ~\ref{Data}, we briefly describe the datasets used in our experiments. We then present details of the pre-trained CNN architecture in Section ~\ref{Network_Architecture}. Various parameters and schemes that influence the recognition rate are described in Section ~\ref{key}. Finally, our comparative evaluation of the performance of the proposed method with that of state-of-the-art methods is provided in Section ~\ref{art}.

\subsection{Dataset} \label{Data}
\begin{figure}
	\centering
	\includegraphics[width=0.96\linewidth]{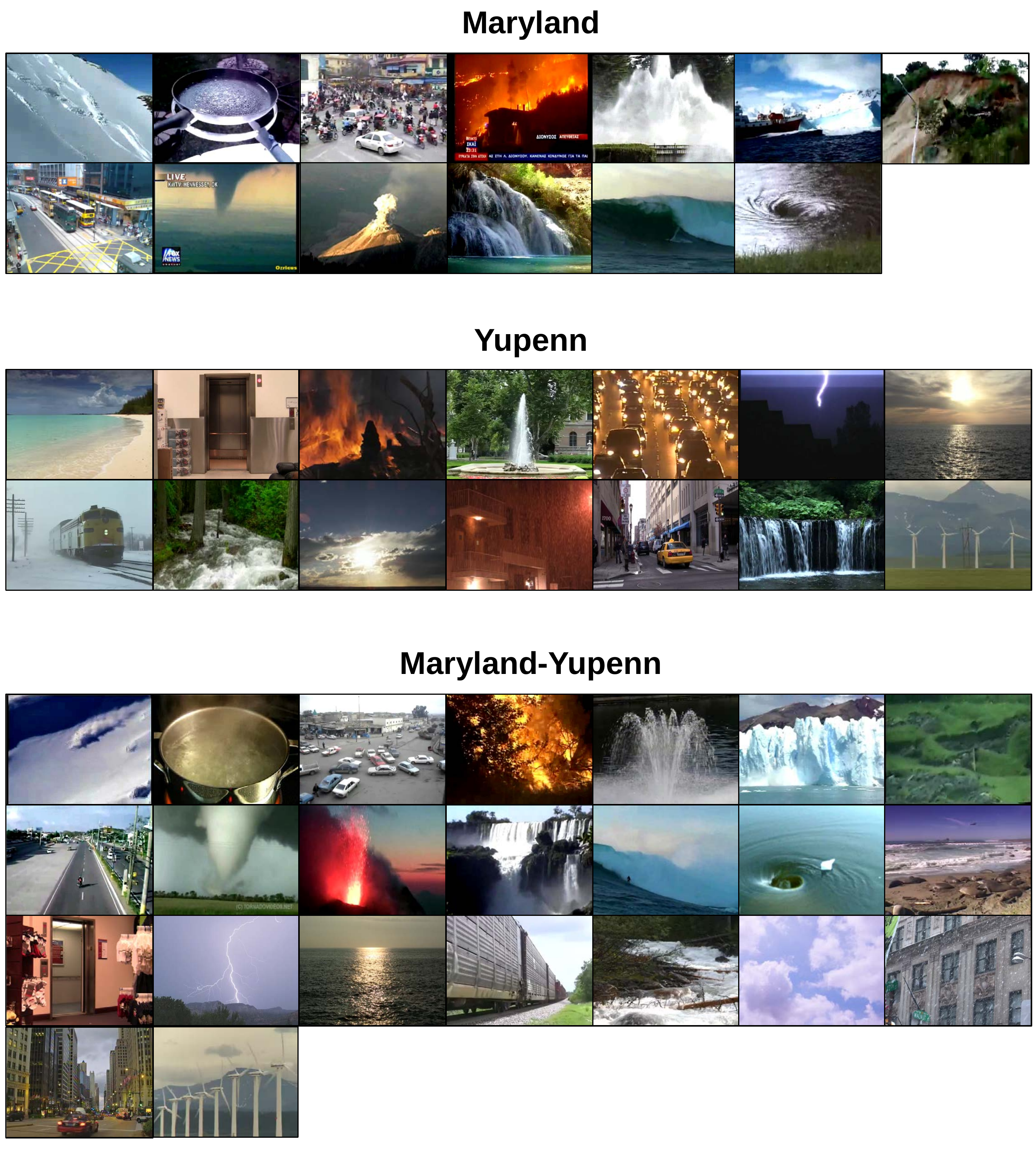}
	\caption{Sample frames from each class in  Maryland, Yupenn, and Maryland-Yupenn datasets}
	\label{fig:MarylandYupenndataset}
\end{figure}

The Maryland dataset \cite{shroff2010moving} contains 13 classes of natural scenes with 10 videos per class. On average, the length of these videos is 617 frames, and the size of each frame is 308$\times$417   pixels. The items in this dataset were collected from various public websites, such as YouTube. Thus, the videos in Maryland have significant camera motion and large variations in illuminations, viewpoints, and scales. Similar to the Maryland dataset, the Yupenn dataset \cite{derpanis2012dynamic} includes variations in viewpoints, scales, and illuminations; however, it does not include camera-introduced motion. This latter dataset consists of 14 dynamic scene classes with 30 videos per class. The average resolution of the videos is  250$\times$370   pixels, and the average frame length is 145. Example frames from the Maryland and Yupenn datasets are shown in Fig.~\ref{fig:MarylandYupenndataset}.

While static scene datasets have been extensively introduced \cite{oliva2001modeling, quattoni2009recognizing, xiao2010sun, cimpoi2015deep, yu2015lsun}, only two public datasets exist for dynamic scenes. To address this lack, we constructed a dataset that includes 23 dynamic scene classes with 10 videos per class by combining the Maryland and Yupenn datasets. Because the number of sequences in each class differs between Maryland (10) and Yupenn (14), we selected 10 random sequences per each class in both datasets. For those classes that exist in both datasets, such as ``Forest Fire,'' ``Fountain,'' and ``Waterfall'', five random sequences were extracted from each dataset and combined. We additionally incorporated ``Smooth Traffic'' in Maryland and ``Highway'' in Yupenn because they have very similar appearances and motion. The whole dataset can be downloaded at our online repository (\href{https://goo.gl/bGPnbu}{https://goo.gl/bGPnbu}).  

\subsection{Pre-trained CNN Architecture} 
\label{Network_Architecture}

\begin{figure}
	\centering
	\includegraphics[width=\linewidth]{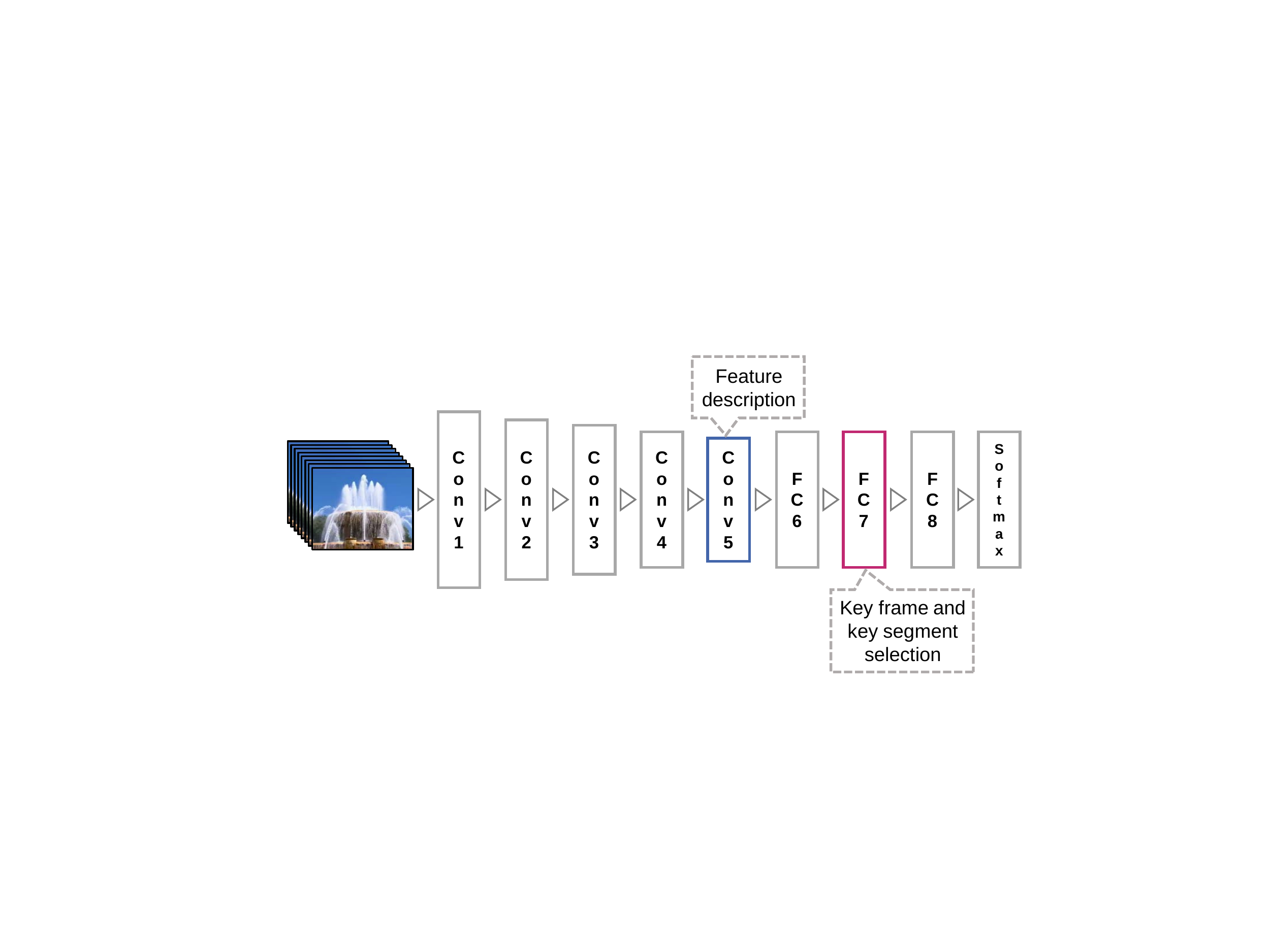}
	\caption{
	Architecture of the VGG-VD model. Among the model layers, the FC7 layer is used to select key frames and key segments, while the Conv5 layer is used to describe the local textures of the key frames and key segments. 	
	}
	\label{fig:cnn}
\end{figure}

Throughout the overall process, we used a pre-trained CNN model, VGG-VD with 16 layers \cite{simonyan2014very}, which was trained on the ImageNet Large Scale Visual Recognition Challenge (ILSVRC) \cite{russakovsky2015imagenet}. This network consisted of five Conv layers containing successive $3\times3$  convolution filters; a stack of Conv layers was followed by three FC layers, as shown in Fig.~\ref{fig:cnn}. Among them, two different types of layers were used in our approach. The FC7 layer was used to select key frames and key segments from a dynamic scene sequence. On the other hand, the Conv5 layer was used to describe local textures of key frames and key segments. The FC7 layer was a 4,096-dimensional vector, which indicates a single global feature with  4,096 dimensions. The output of the Conv5 layer was  7$\times$7$\times$512, which yielded  49 (7$\times$7) local features with 512  dimensions for each frame. For further details, please refer to \cite{simonyan2014very}.

\subsection{Key Factors} \label{key}
In this section, we present a comparative analysis of the various factors---the frame selection strategy, number of frames, layer for feature description, codebook size, and combining of spatial and temporal features---that influence the performance of the proposed method. Analytical experiments were performed on the challenging Maryland dataset, following the leave-one-out cross-validation approach described in \cite{shroff2010moving}. For classification, a support vector machine (SVM) \cite{cortes1995support}, where parameter C was fixed to 100, was used. We additionally evaluated key factors in the Maryland--Yupenn dataset following the experimental setup in the Maryland dataset.

\subsubsection{Frame Selection Strategy}  
When processing the spatial features from a video in a frame-by-frame manner, a large difference of performance could be caused with respect to frame selection strategy. We compared the proposed method with previous approaches on the same setting. For a fair comparison, we used only static features from the key frames, i.e.,  D3$_s$ with FV aggregation wherein codebook size was 128.

The baseline frame selection strategies were divided into two types: fixed frame number and varying frame number. In the comparative experiments for the fixed frame number setting, we evaluated the random sampling introduced in \cite{yang2016dynamic}. We furthermore applied adaptive key frame extraction based on clustering \cite{zhuang1998adaptive} (CHC in the table). In this approach, all the frames were grouped by a  $k$-means algorithm, and the closest frame to the cluster centroid was selected as the key frame.

The comparison results with the fixed frame number approaches are presented in Table~\ref{table:fixed_frameNumber}. From the table, the tendency of performance improvement is evident with the increase of the number of frames in most cases. Interestingly, the CHC approach outperforms the random sampling except when only one frame is used. This result indicates that one key frame, which was obtained by  $k$-means clustering, may not well represent a dynamic scene sequence. In contrast, the proposed method shows a relatively promising result with only one frame. Overall, the results from the Maryland dataset (Table~\ref{table:fixed_frameNumber}) and those from the Maryland–Yupenn dataset (Table~\ref{table:fixed_frameNumber_MY}) support the superiority of the proposed method.

\begin{table}
	\renewcommand{\arraystretch}{1.3}
	\caption{Classification rate (\%) of the methods using a fixed frame number on the Maryland dataset}
	\label{table:fixed_frameNumber}
	\centering		
	\begin{small}
		\begin{tabular}{p{16mm}cccc}		
			\toprule[1pt]
			\multirow{2}{*}{\bf{Selection}}   		 
			&\multicolumn{4}{c}{\bf{Frame Number}} \\ 				
			& 1 &  5	& 10 & 15		\\ \hline					
			Random  & 73.08 &	80.77 	&85.38 &	86.15	\\ 		
			CHC	&70.77  &	84.62 &	86.92 &	84.62      \\ 
			\boldmath {D3$_s$}  	&\textbf{75.38} &	\textbf{84.62} &	\textbf{89.23} &	\textbf{90.77}      \\		  
			\bottomrule[1pt]			   
		\end{tabular}
	\end{small}	
\end{table}

\begin{table}
	\renewcommand{\arraystretch}{1.3}
	\caption{Classification rate (\%) of the methods using a fixed frame number on the Maryland--Yupenn dataset}
	\label{table:fixed_frameNumber_MY}
	\centering		
	\begin{small}
		\begin{tabular}{p{16mm}cccc}		
			\toprule[1pt]
			\multirow{2}{*}{\bf{Selection}}   		 
			&\multicolumn{4}{c}{\bf{Frame Number}} \\ 				
			& 1 &  5	& 10 & 15	 	\\ \hline					
			Random  & 78.70 &	88.26 &	86.09 	&87.39 	\\ 		
			CHC	&  81.30 &	89.13 	&89.57 &	89.57  \\ 
			\boldmath {D3$_s$}  &	\textbf{82.17} &	\textbf{90.87} &	\textbf{91.74} &	\textbf{91.74}           \\ 		  
			\bottomrule[1pt]			 
		\end{tabular}
	\end{small}	
\end{table}

\begin{table}
	\renewcommand{\arraystretch}{1.3}
	\caption{
	Comparison results of the methods using a varying frame number on the Maryland dataset}
	\label{table:varing_frameNumber}
	\centering		
	\begin{small}
		\begin{tabular}{p{18mm}cc}		
			\toprule[1pt]
			\bf{Selection} & \bf{Parameter} & \bf{Accuracy} (\%) \\ \hline		
			\multirow{5}{*}{Consecutive} & 1 & 74.62   \\ 	
			&N/8	  & 80.77 \\ 
			&	N/4	  & 83.08 \\ 
			&	N/2	  & 84.62 \\ 
			&	N	  & 88.46 \\ 	\hline		
			\multirow{3}{*}{GHD} & $\mu - \sigma$ & 86.15    \\ 	
			& $\mu$ &87.69 			    \\ 		
			& $\mu + \sigma$ &  87.69   \\ 		\hline		
			
			Thumbnail & 10 s & 76.15 \\ 			\hline					
			\boldmath {D3$_s$}  & 15 frames & \textbf{90.77} \\  
			\bottomrule[1pt]			  
		\end{tabular}
	\end{small}	
\end{table}

In addition, we compared the proposed approach with several frame selection approaches based on a varying frame number. The results from the Maryland dataset are presented in Table~\ref{table:varing_frameNumber}. As a baseline, we first evaluated a partially consecutive sampling of \cite{qi2016dynamic} (Consecutive in the table). A second approach is a histogram-difference-based approach  (GHD in the table). The histogram differences between adjacent gray images were obtained for all frames and their mean and standard deviations were used as a threshold to remove outlier frames. One standard deviation subtracted from the mean($\mu - \sigma$), the mean ($\mu$), and one standard deviation plus the mean ($\mu + \sigma$) were used in our experiments. On average, the number of frames was 3.76\%, 68.79\%, and 90.59\% of the total number of frames, respectively. We additionally applied video-thumbnail extraction of FFMPEG, which extracted the most representative frame per 10 s (Thumbnail in the table). From the table, it is apparent that the proposed method outperformed the conventional sampling strategies and key frame selection.

\subsubsection{Description Layer and Codebook Size}  
While features via the FC layer well describe the overall structure of the scene, the features via the Conv layer that we used considered the local textures of the scene. In this section, we compare the results between the FC layer and Conv layer in terms of the description layer. In both cases, key frame selection was performed by the FC layer, and FV was used for feature aggregation. The performance against codebook size used in FV aggregation was also given in the experiments. We used the $k$-means++ algorithm \cite{arthur2007k} for the codebook generation, where the termination condition occurred after 1,000 iterations or when the positions of new centroids did not change in the new iteration.

From Table~\ref{table:codebook}, it is clear that the Conv layer outperforms that of the FC layer with respect to the number of frames and codebook size. It is evident that a different tendency exists with the increase of codebook size. The performance gradually decreases when using the FC layer, whereas the performance increases when using the Conv layer. The best performance in this experiment is 90.77\% when the number of frames is 15 and the codebook size is 128.

\begin{table}
	\renewcommand{\arraystretch}{1.3}
	\caption{
	Classification rate (\%) of D3$_s$  with respect to the number of frames, feature description layer, and codebook size on the Maryland dataset}
	\label{table:codebook}
	\centering		
	\begin{small}
		\begin{tabular}{ccccccc}		
			\toprule[1pt]
			 \multirow{2}{*}{\bf{Frame}}  &  \multirow{2}{*}{\bf{Layer}} 	&\multicolumn{5}{c}{\bf{Codebook Size}}		\\	
			  &  & 32 & 64 & 128&	256&	512 \\ \hline
			  \multirow{2}{*}{\bf{1}} & FC &	66.15 &	69.23 &	68.46 &	68.46 &	68.46 \\ 
			    & Conv&	73.85 &	74.62 	&75.38&	76.92 &	\bf{80.00} \\ 
			    \hline
			  \multirow{2}{*}{\bf{5}}  & FC&	81.54 &	82.31 &	75.38 	&72.31 &	66.92 \\ 
			    & Conv&	82.31& 	82.31 &	84.62 	&86.15 	&\bf{89.23} \\ 
			      \hline
			\multirow{2}{*}{\bf{10}}  & FC	&82.31 &	80.00& 	82.31 &76.92 	&72.31 \\ 
			  &  Conv&	84.62 &	86.15 &	\bf{89.23} &	\bf{89.23}& 	\bf{89.23} 			  \\ 
			   \hline
			\multirow{2}{*}{\bf{15}} & FC&	83.85 &	79.23 &	82.31 &	69.23 &	76.15 \\ 
			 & Conv&	82.31 &	83.85 &	\bf{90.77} &	90.00 	&88.46 \\ 			 
		    \bottomrule[1pt]			   
		\end{tabular}
	\end{small}	
\end{table}

\subsubsection{Spatial and Temporal Features}

\begin{figure}
	\centering
	\includegraphics[width=1.01\linewidth]{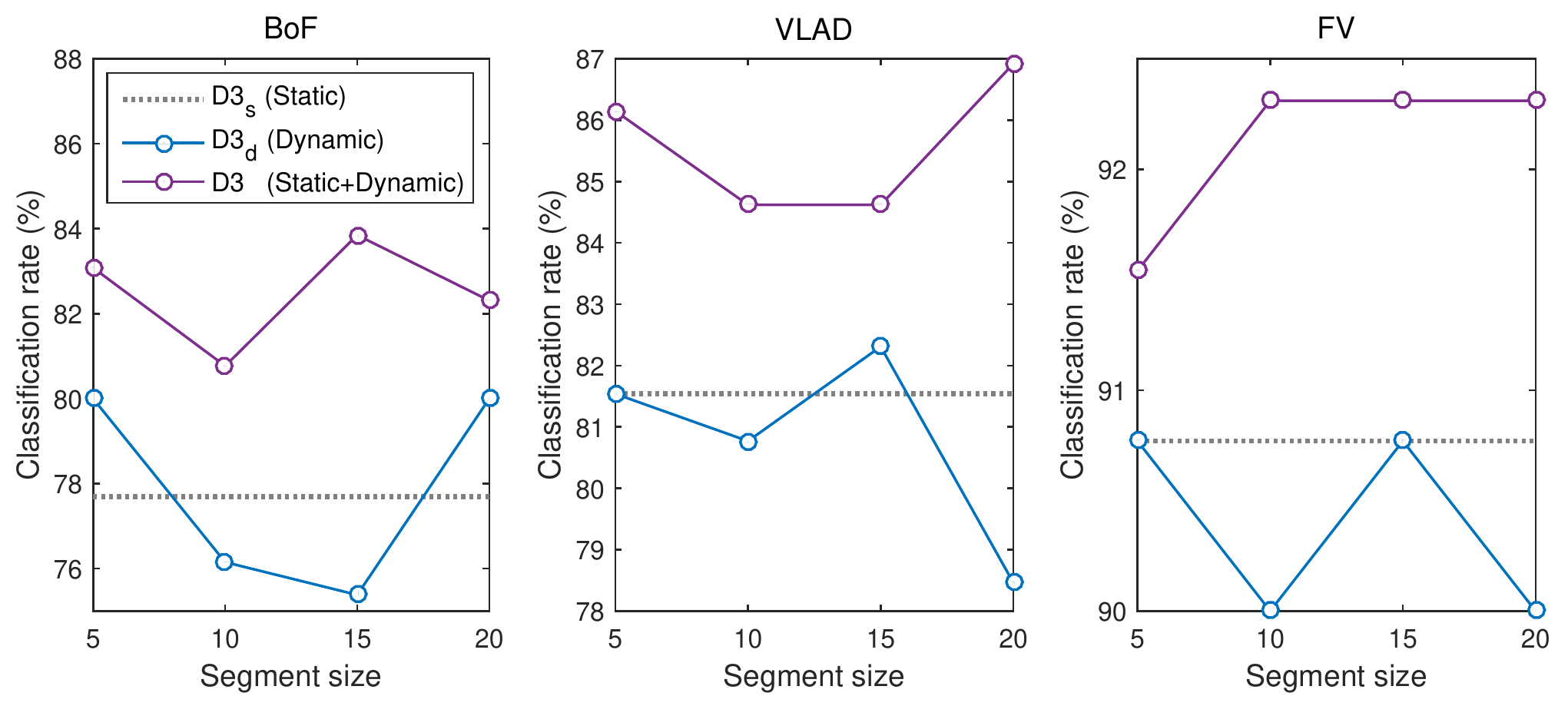}
	\caption{Classification rate of the individual descriptor using different feature aggregations and segment sizes on the Maryland dataset.}
	\label{fig:static_dynamic}
\end{figure}

To describe a dynamic scene, we calculated the static appearance of key frames,  D3$_s$, and dynamic patterns in key segments,  D3$_d$. We then combined D3$_s$  and D3$_d$  for video descriptor D3. Fig.~\ref{fig:static_dynamic} shows the performance of individual descriptors using BoF, VLAD, and FV feature aggregation. The baseline (represented by a dotted horizontal line) is the result of using the static feature alone. We can see that the combined feature gives better performance than using the static feature or dynamic feature alone. In this experiment, we confirmed 92.31\% of  D3 with FV, which is the state-of-the-art performance in the Maryland dataset.

\subsection{Comparison with State-of-the-Art Methods} \label{art}  
In this section, we compare the proposed method with state-of-the-art methods on the public dynamic scene datasets. We additionally evaluate the proposed method on the newly compiled Maryland--Yupenn dataset. Finally, we show comparison results on dynamic texture datasets to demonstrate the generalization ability of the proposed method.

\subsubsection{Dynamic Scene Dataset} 
For a fair comparison with previous studies on Maryland and Yupenn, we followed the leave-one-out evaluation protocol and reported classification results using SVM. Table~\ref{table:RESULTS_Maryland_Yupenn} shows the comparison results, which can be divided into hand-crafted features \cite{theriault2013dynamic, feichtenhofer2013spacetime, feichtenhofer2014bags, yang2016dynamic, feichtenhofer2016dynamic} and a CNN feature based on deep learning \cite{jia2014caffe, tran2015learning, qi2016dynamic}. In the experiments, several hand-crafted features, such as bags of spacetime energies (BoSE), ensemble SVM (E-SVM), and dynamically pooled complementary features (DPCF), show good performance in the stabilized Yupenn dataset. However, they show poor performance compared to the CNN-based methods, including ours, in the non-stabilized Maryland dataset.

\begin{table}
	\renewcommand{\arraystretch}{1.3}
	\caption{
	Comparison results using different methods in the Maryland and Yupenn datasets}
	\label{table:RESULTS_Maryland_Yupenn}
	\centering		
	\begin{small}
		\begin{tabular}{p{20mm}cc}		
			\toprule[1pt]
			 \multirow{2}{*} {\head{Method}}  & \multicolumn{2}{c} {\head{Accuracy (\%)}} \\ 
			  & Maryland & Yupenn  \\ \hline			 
	  	   SFA \cite{theriault2013dynamic}	& 74.60 &	85.00     \\		
			CSO	\cite{feichtenhofer2013spacetime}		&	67.69 &	85.95			 \\ 
			BoSE \cite{feichtenhofer2014bags}		&	77.69 &	96.19			 \\ 			
			E-SVM \cite{yang2016dynamic}		&	78.77 &	96.43			 \\ 	
			DPCF \cite{feichtenhofer2016dynamic}		 & 80.00 &	98.81 		   \\ 	
			Imagenet \cite{jia2014caffe}  & 87.70 &	96.70		   \\ 	
			C3D	\cite{tran2015learning}  &87.70 &	98.10		  \\ 
			st-TCoF \cite{qi2016dynamic} & 88.46 & \bf{99.05}      \\			
			\boldmath {D3}     &	\bf{92.31}	&   \bf{99.05} 		 \\ 
				\bottomrule[1pt]	
		\end{tabular}
	\end{small}	
\end{table}

\begin{table}
	\renewcommand{\arraystretch}{1.3}
	\caption{
	Comparison results between TcoF-based approaches and the proposed method in the Maryland and Yupenn datasets
	}
	\label{table:RESULTS_TCOF_D3}
	\centering		
	\begin{small}
		\begin{tabular}{p{14mm}p{20mm}cc}		
			\toprule[1pt]
			& \multirow{2}{*} {\head{Method}}  & \multicolumn{2}{c} {\head{Accuracy (\%)}} \\ 
		   & 	  & Maryland & Yupenn  \\ \hline				
			\multirow{2}{*} {Static} &	s-TCoF \cite{qi2016dynamic} & 88.46 & 97.14  \\ 
					 & \boldmath {D3$_s$}     &	\bf{90.77} 	&  \bf{97.62}  \\  \hline
			\multirow{2}{*} {Dynamic} &		t-TCoF \cite{qi2016dynamic}& 66.15 &  97.86	 \\ 
					& \boldmath {D3$_d$}     &	\bf{90.77} 		& \bf{98.33} 	\\  \hline			
		    Static \texttt{+} & 	st-TCoF \cite{qi2016dynamic} & 88.46 & \bf{99.05}      \\			
			Dynamic  & \boldmath {D3}     &	\bf{92.31}	&   \bf{99.05} 		 \\ 
				\bottomrule[1pt]	
		\end{tabular}
	\end{small}	
\end{table}

In considering separate modeling of dynamic scenes and the transferred CNN model, TCoF methods, including s-TCoF, t-TCoF, and st-TCoF, are most closely related to our  D3$_s$,  D3$_d$, and D3  respectively. We performed direct comparative experiments of TCoF methods and our approach. As shown in Table~\ref{table:RESULTS_TCOF_D3}, TCoF methods show promising results except for t-TCoF in Maryland dataset. This result may be because of their temporal encoding based on differences between two adjacent frames, which motivated this study. In contrast, our  D3$_d$ based on key segments shows good performance. We can also see dynamic feature  D3$_d$  provides complementary information when combined with static feature  D3$_s$. Overall,  D3 shows the best result among the methods in both Maryland and Yupenn datasets.

\begin{table}
	\renewcommand{\arraystretch}{1.3}
	\caption{
	Classification rate (\%) with respect to different methods and feature aggregation on the Maryland--Yupenn dataset}
	\label{table:RESULTS_MARYLANDYUPENN}
	\centering		
	\begin{small}
		\begin{tabular}{p{14mm}ccc}		
			\toprule[1pt]
		   \multirow{2}{*} {\head{Method}}  & \multicolumn{3}{c} {Feature aggregation} \\ 
		   & 	BoF  & VLAD & FV  \\ \hline	
		   	  \boldmath {D3$_s$}     &	82.61 	&  81.74 & 	\bf{90.43} \\
			 \boldmath {D3$_d$}     &	80.87		& 84.78 & 	\bf{91.30} 	\\  	
		 	 \boldmath {D3}     &86.96	&   87.83	& 	\bf{92.61}	 \\ 
				\bottomrule[1pt]	
		\end{tabular}
	\end{small}	
\end{table}

The results from the Maryland--Yupenn dataset are also presented in Table~\ref{table:RESULTS_MARYLANDYUPENN}. We followed an experimental setup similar to the one in the Maryland and Yupenn datasets. As expected,  D3 outperformed   D3$_s$ and D3$_d$  with all feature aggregation approaches. The best results of each descriptor are   D3$_s$ (90.43\%),  D3$_d$ (91.30\%), and  D3 (92.61\%), respectively, which can be baselines in further experiments in the Maryland--Yupenn dataset.

\subsubsection{Dynamic Texture Dataset}

\begin{figure}[h]
	\centering
	\includegraphics[width=0.85\linewidth]{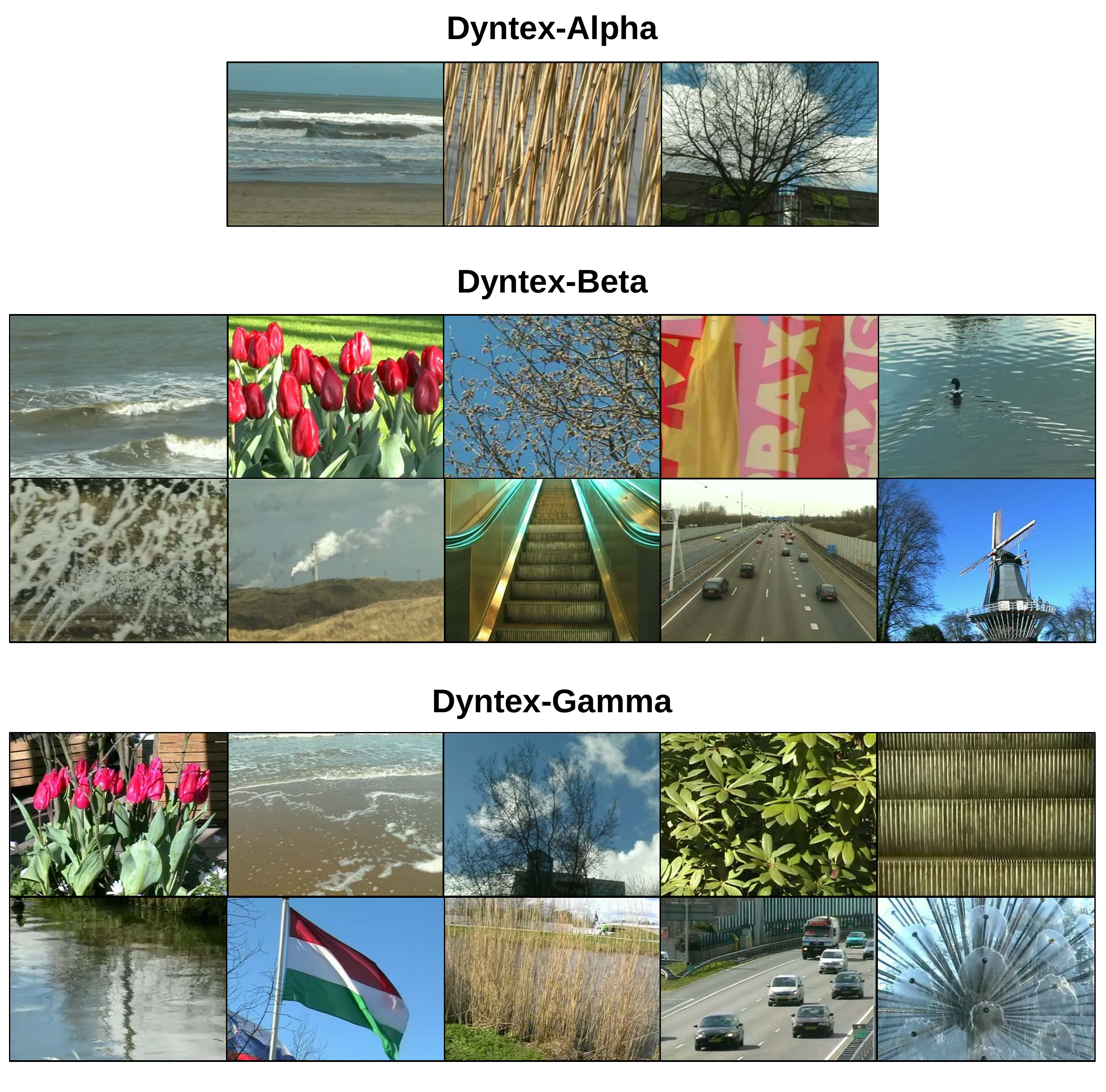}
	\caption{Sample frames from each class in Dyntex-Alpha, Dyntex-Beta, and Dyntex-Gamma datasets}
	\label{fig:Dyntex_new}
\end{figure}

To validate the effectiveness of the proposed method, we also evaluated the experiments on dynamic textures datasets: Dyntex-Alpha, Dyntex-Beta, and Dyntex-Gamma \cite{dubois2015characterization}. The Dyntex-Alpha dataset is composed of 60 DT sequences that are equally divided into three categories, and the Dyntex-Beta consists of 162 sequences with 10 classes. The Dyntex-Gamma, which is the most challenging set, is composed of 264 sequences grouped into 10 classes. Sample frames of each dataset are given in Fig.~\ref{fig:Dyntex_new}. Table~\ref{table:RESULTS_DYNTEX-New} shows overall comparison results in each dynamic texture dataset. We can see that the proposed  D3  outperforms the previous works in all dynamic texture datasets.

\begin{table}[h]
	\renewcommand{\arraystretch}{1.3}
	\caption{Comparison results using different methods with Dyntex-Alpha, Dyntex-Beta, and Dyntex-Gamma datasets}
	\label{table:RESULTS_DYNTEX-New}
	\centering		
	\begin{small}
		\begin{tabular}{p{17mm}ccc}		
					\toprule[1pt]
					 \multirow{2}{*} {\head{Method}}  & \multicolumn{3}{c} {\head{Accuracy(\%)}} \\ 
							  & Dyntex-Alpha & Dyntex-Beta  &  Dyntex-Gamma\\  \hline			 			
					OTF \cite{xu2012scale}	& 	 	83.61 & 	 	73.22 & 	 	72.53 \\
					WMF	\cite{ji2013wavelet} & 	 	84.83 	&  	75.21 &	 	73.32 \\
					DFS	\cite{xu2015classifying} &	 	85.24 & 	 	76.93 & 		74.82 \\
					LA \cite{sun2015characterizing} &	 	89.60 & 	 	80.90 	&	79.90 \\
					s-TCoF \cite{qi2016dynamic}  & 100.00  &  \bf{100.00}	&   \bf{98.11} \\ 
						t-TCoF \cite{qi2016dynamic} & 98.33 & 97.53 	&95.45 \\ 
						st-TCoF \cite{qi2016dynamic} &  \bf{100.00}&\bf{100.00}&\bf{98.11}\\ 													
					\boldmath {D3$_s$}     &\bf{100.00}    &  98.77 &	97.35\\ 
					\boldmath {D3$_d$}     & \bf{100.00}   &  98.77 & 	\bf{98.11} 	\\ 
					\boldmath {D3}     &  \bf{100.00}  &  \bf{100.00} 	& \bf{98.11} \\ 						
					\bottomrule[1pt]								    
		\end{tabular}
	\end{small}	
\end{table}
   \vspace*{-0.1in}

\section{Conclusion}
In this study, we have proposed a new dynamic scene representation based on `key frame' and `key segment.' Given a sequence, considering feature distribution obtained by FC layer, we first select key frames and key segments. With Conv layer, static features are then extracted from key frames while dynamic features are calculated from key segments. Finally, static and dynamic features are aggregated by FV separately, followed by late fusion, which constructs a deep dual descriptor (D3). We validated our approach by outperforming the conventional frame selection strategies. From the experiments, we also confirmed that our D3 using Conv layer outperforms that of FC layer. Results using three dynamic scene datasets and three dynamic texture datasets further demonstrate the superiority of the proposed method compared to state-of-the-art methods. In our approach, we utilize only part of the deep learning architecture by using transferred CNN feature with conventional feature aggregation. We plan to expand our approach to full CNN architecture in which the feature aggregation and decision were jointly performed on the CNN architecture.

\bibliographystyle{IEEEbib}
\bibliography{refs}

\end{document}